\pdfoutput=1

\documentclass[11pt]{article}

\usepackage[table]{xcolor}
\usepackage[final]{acl}

\usepackage{times}
\usepackage{latexsym}

\usepackage[T1]{fontenc}

\usepackage[utf8]{inputenc}

\usepackage{microtype}

\usepackage{inconsolata}

\usepackage{graphicx}

\usepackage{booktabs}
\usepackage{xspace}
\usepackage{microtype}
\usepackage{amsfonts}
\usepackage{amsmath}
\usepackage{amssymb}
\usepackage{amsthm}
\usepackage{booktabs}
\usepackage{dsfont}
\usepackage{paralist}
\usepackage{times}
\usepackage{xspace}
\usepackage{multirow}
\usepackage{graphicx}
\usepackage{caption}
\usepackage{subcaption}
\usepackage{float} 
\usepackage{lipsum}
\usepackage{appendix}
\usepackage{pifont}
\usepackage{makecell}
\usepackage{caption}
\usepackage{latexsym}
\usepackage{cleveref}
\usepackage{subcaption}
\usepackage{placeins}
\usepackage{upgreek}
\usepackage[inline]{enumitem}

\usepackage[export]{adjustbox}

\crefformat{section}{\S#2#1#3}
\crefformat{figure}{Fig.~#2#1#3}
\crefformat{table}{Tab.~#2#1#3}
\crefformat{appendix}{\S#2#1#3}

\renewcommand\paragraph[1]{
\vspace{0.15cm}
\noindent 
\textbf{#1}
}

\newcommand{\ra}[1]{\renewcommand{\arraystretch}{#1}}
\definecolor{lightgray}{gray}{0.95}
\usepackage{soul}

\usepackage{amsmath}
\usepackage{textcomp}
\usepackage{gensymb}

\usepackage[edges]{forest}
\usepackage[most]{tcolorbox}
\usepackage[tikz]{bclogo}
\usepackage[framemethod=tikz]{mdframed}
\definecolor{bgblue}{RGB}{245,243,253}
\definecolor{ttblue}{RGB}{91,194,224}

\tikzset{%
    parent/.style =          {align=center,text width=3cm, rounded corners=3pt, line width=0.3mm, fill=gray!10,draw=gray!80},
    child/.style =           {align=center,text width=2.3cm,rounded corners=3pt, fill=blue!10,draw=blue!80,line width=0.3mm},
    grandchild/.style =      {align=center,text width=2cm,rounded corners=3pt},
    greatgrandchild/.style = {align=center,text width=1.5cm,rounded corners=3pt},
    greatgrandchild2/.style = {align=center,text width=1.5cm,rounded corners=3pt},    
    referenceblock/.style =  {align=center,text width=1.5cm,rounded corners=2pt},
    data_based/.style =           {align=center,text width=3cm,rounded corners=3pt, fill=blue!10,draw=blue!80,line width=0.3mm},   
    data_based_work/.style =           {align=center, text width=6cm,rounded corners=3pt, fill=blue!10,draw=blue!0,line width=0.3mm},  
    model_based/.style =           {align=center,text width=3cm,rounded corners=3pt, fill= orange!10,draw= orange!80,line width=0.3mm},   
    model_based_work/.style =           {align=center,text width=6cm, rounded corners=3pt, fill= orange!10,draw= orange!0,line width=0.3mm},            
}

\title{Data Contamination Report from the 2024 CONDA Shared Task}

\author{
 \textbf{Oscar Sainz\textsuperscript{1}}\quad
 \textbf{Iker Garc\'ia-Ferrero\textsuperscript{1}}\quad
 \textbf{Alon Jacovi\textsuperscript{2}}
\\
 \textbf{Jon Ander Campos\textsuperscript{3}}\quad
 \textbf{Yanai Elazar\textsuperscript{4,5}}\quad
 \textbf{Eneko Agirre\textsuperscript{1}}\quad
 \textbf{Yoav Goldberg\textsuperscript{2,4}}
\\ \\
 \textbf{Wei-Lin Chen\textsuperscript{6,7}}\quad
 \textbf{Jenny Chim\textsuperscript{8}}\quad
 \textbf{Leshem Choshen\textsuperscript{9,10}}\quad
 \textbf{Luca D'Amico-Wong\textsuperscript{11}}\\
 \textbf{Melissa Dell\textsuperscript{11}}\quad
 \textbf{Run-Ze Fan\textsuperscript{12}}\quad
 \textbf{Shahriar Golchin\textsuperscript{13}}\quad
 \textbf{Yucheng Li\textsuperscript{14}}\quad
 \textbf{Pengfei Liu\textsuperscript{12}}\\
 \textbf{Bhavish Pahwa\textsuperscript{15}}\quad
 \textbf{Ameya Prabhu\textsuperscript{16,17}}\quad
 \textbf{Suryansh Sharma\textsuperscript{18}}\quad
 \textbf{Emily Silcock\textsuperscript{11}}\\
 \textbf{Kateryna Solonko\textsuperscript{15}}\quad
 \textbf{David Stap\textsuperscript{19}}\quad
 \textbf{Mihai Surdeanu\textsuperscript{13}}\quad
 \textbf{Yu-Min Tseng\textsuperscript{6}}\\
 \textbf{Vishaal Udandarao\textsuperscript{17,21}}\quad
 \textbf{Zengzhi Wang\textsuperscript{12}}\quad
 \textbf{Ruijie Xu\textsuperscript{12}}\quad
 \textbf{Jinglin Yang\textsuperscript{11}}
\\
\\
 \textsuperscript{1}HiTZ Center - Ixa, University of the Basque Country UPV/EHU\;
 \textsuperscript{2}Bar Ilan University\;
 \textsuperscript{3}Cohere\\
 \textsuperscript{4}Allen Institute for Artificial Intelligence\;
 \textsuperscript{5}University of Washington\;
 \textsuperscript{6}National Taiwan University\\
 \textsuperscript{7}University of Virginia\;
 \textsuperscript{8}Queen Mary University of London\;
 \textsuperscript{9}MIT-IBM Watson AI Lab\;
 \textsuperscript{10}MIT\\
 \textsuperscript{11}Harvard University\;
 \textsuperscript{12}Shanghai Jiao Tong University\;
 \textsuperscript{13}University of Arizona\;
 \textsuperscript{14}University of Surrey\\
 \textsuperscript{15}Microsoft Research\;
 \textsuperscript{16}Tübingen AI Center\;
 \textsuperscript{17}University of Tübingen\\
 \textsuperscript{18}Indian Institute of Technology Kharagpur\;
 \textsuperscript{19}University of Amsterdam\;
 \textsuperscript{20}Microsoft\;
 \textsuperscript{21}University of Cambridge\;
\\
 \small{
   \texttt{Contact: \href{mailto:conda-workshop@googlegroups.com}{conda-workshop@googlegroups.com}}
 }
 }

\begin{document}
\maketitle

\begin{abstract}

    The 1st Workshop on Data Contamination (CONDA 2024) focuses on all relevant aspects of data contamination in natural language processing, where data contamination is understood as situations where evaluation data is included in pre-training corpora used to train large scale models, compromising evaluation results. The workshop fostered a shared task to collect evidence on data contamination in current available datasets and models. The goal of the shared task and associated database is to assist the community in understanding the extent of the problem and to assist researchers in avoiding reporting evaluation results on known contaminated resources. The shared task provides a structured, centralized public database for the collection of contamination evidence, open to contributions from the community via GitHub pool requests. This first compilation paper is based on 566 reported entries over 91 contaminated sources from a total of 23 contributors. The details of the individual contamination events are available in the platform.\textsuperscript{1} The platform continues to be online, open to contributions from the community.
\end{abstract}

\section{Introduction}

\begin{figure*}[!ht]
    \centering
    \resizebox{.8\textwidth}{!}{
        \begin{forest}
            for tree={
            forked edges,
            grow'=0,
            draw,
            rounded corners,
            node options={align=center,},
            text width=3cm,
            s sep=6pt,
            calign=child edge, calign child=(n_children()+1)/2,
            },
            [Contamination detection methods, fill=gray!45, parent
            [Data-based, for tree={ data_based}
            [Proprietary data,  data_based
            [GPT-3~\cite{brown2020language}, style = data_based_work]
            [FLAN~\cite{wei2022finetuned}, style = data_based_work]
            [GLaM~\cite{du2022glam}, style = data_based_work]
            [PaLM~\cite{chowdhery2022palm}, style = data_based_work]
            [PaLM-2~\cite{anil2023palm}, style = data_based_work]
            [GPT-4~\cite{openai2024gpt4}, style = data_based_work]
            ]
            [Open data, data_based
            [~\citet{dodge-etal-2021-documenting}, style = data_based_work]
            [~\citet{silcock2023noiserobust}, style = data_based_work]
            [~\citet{muennighoff-etal-2023-crosslingual}, style = data_based_work]
            [~\citet{azerbayev2023llemma}, style = data_based_work]
            [~\citet{elazar2024whats}, style = data_based_work]
            [~\citet{riddell2024quantifying}, style = data_based_work]
            [~\citet{li2024open}, style = data_based_work]
            ]
            ]
            [Model-based, for tree={model_based}
            [Closed model, model_based
            [~\citet{sainz2023chatgpt}, style = model_based_work]
            [~\citet{golchin2024time}, style = model_based_work]
            [~\citet{golchin2024data}, style = model_based_work]
            [~\citet{dong2024generalization}, style = model_based_work]
            [~\citet{ranaldi2024investigating}, style = model_based_work]
            [~\citet{enis2024llm}, style = model_based_work]
            ]
            [Open model, model_based
            [~\citet{deng-etal-2024-investigating}, style = model_based_work]
            [~\citet{oren2024proving}, style = model_based_work]
            [~\citet{xu2024benchmarking}, style = model_based_work]
            ]
            ]
            ]
        \end{forest}
    }
    \caption{Taxonomy of papers that report contamination evidence. Including LLM's papers and technical reports, papers about methods for detecting contamination, and papers about corpus analysis.}
    \label{fig:typo-methods}
\end{figure*}
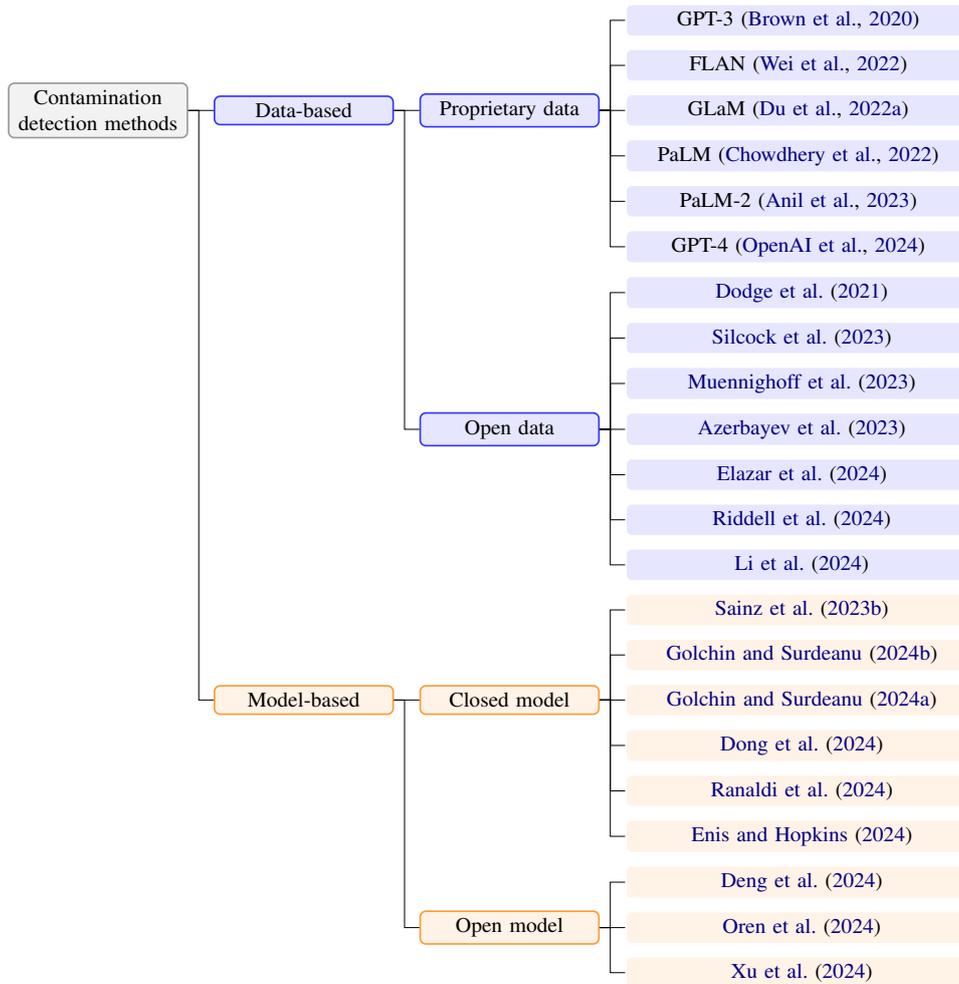

Data contamination, where evaluation data is inadvertently included in pre-training corpora of large-scale models, and language models (LMs) in particular, has become a concern in recent times \cite{sainz-etal-2023-nlp,jacovi-etal-2023-stop}. The growing scale of both models and data, coupled with massive web crawling, has led to the inclusion of segments from evaluation benchmarks in the pre-training data of LMs \cite{dodge-etal-2021-documenting,openai2024gpt4,anil2023palm,elazar2024whats}. The scale of internet data makes it difficult to prevent this contamination from happening, or even detect when it has happened  \cite{bommasani2022opportunities,mitchell2023measuring}. 

Crucially, when evaluation data becomes part of pre-training data, it introduces biases and can artificially inflate the performance of LMs on specific tasks or benchmarks \citep{magar-schwartz-2022-data,Magnusson2023PalomaAB,ngram-novelty}. This poses a challenge for fair and unbiased evaluation of models, as their performance may not accurately reflect their generalization capabilities \citep{Hupkes2023}. And similarly to pre-training contamination, the contamination can also occur during the fine-tuning stage even \textit{after} a model has been deployed as an API \cite{balloccu-etal-2024-leak}.

Although a growing number of papers and state-of-the-art models mention issues of data contamination \cite{brown2020language,wei2022finetuned,chowdhery2022palm,openai2024gpt4,anil2023palm,touvron2023llama}, there is little in the way of organized and compiled knowledge about real, documented cases of contamination in practice \cite{sainz-etal-2023-nlp}. Addressing data contamination is a shared responsibility among researchers, developers, and the broader community.

This report compiles the evidence reported in the Data Contamination Database\footnote{\url{https://huggingface.co/spaces/CONDA-Workshop/Data-Contamination-Database}} as part of the Data Contamination Workshop.\footnote{\url{https://conda-workshop.github.io/}} As the Shared Task of the workshop, researchers were invited to discover cases of contamination in available corpora and models, and submit evidence of their discovery. The submissions to the database were collected and compiled on June 23rd, 2024, to be included in this report, but the database continues to run and grow. Overall we collected 566 submissions from 23 contributors, where each submission included a detailed contamination report, indicating the estimated percentage of contaminated data. 
We continue to operate the database, and expect to update it with newer datasets and models as they come out, as well as new report about existing contaminated (or uncontaminated) evaluations.

This report first presents the methodology for collecting evidence, as well as existing papers that report data contamination (Section \ref{sec:methodology}). We also report the evidence collected in the Data Contamination Database (Section \ref{sec:compilation}), followed by an overview of the trends and statistics in the database, that inform a high-level perspective on the state of data contamination in NLP today (Section \ref{sec:trends}). 

\section{Methodology and Previous Work}
\label{sec:methodology}

\begin{figure*}[t]
    \centering
    \begin{minipage}[t]{0.48\linewidth}
        \centering
        \includegraphics[width=\linewidth]{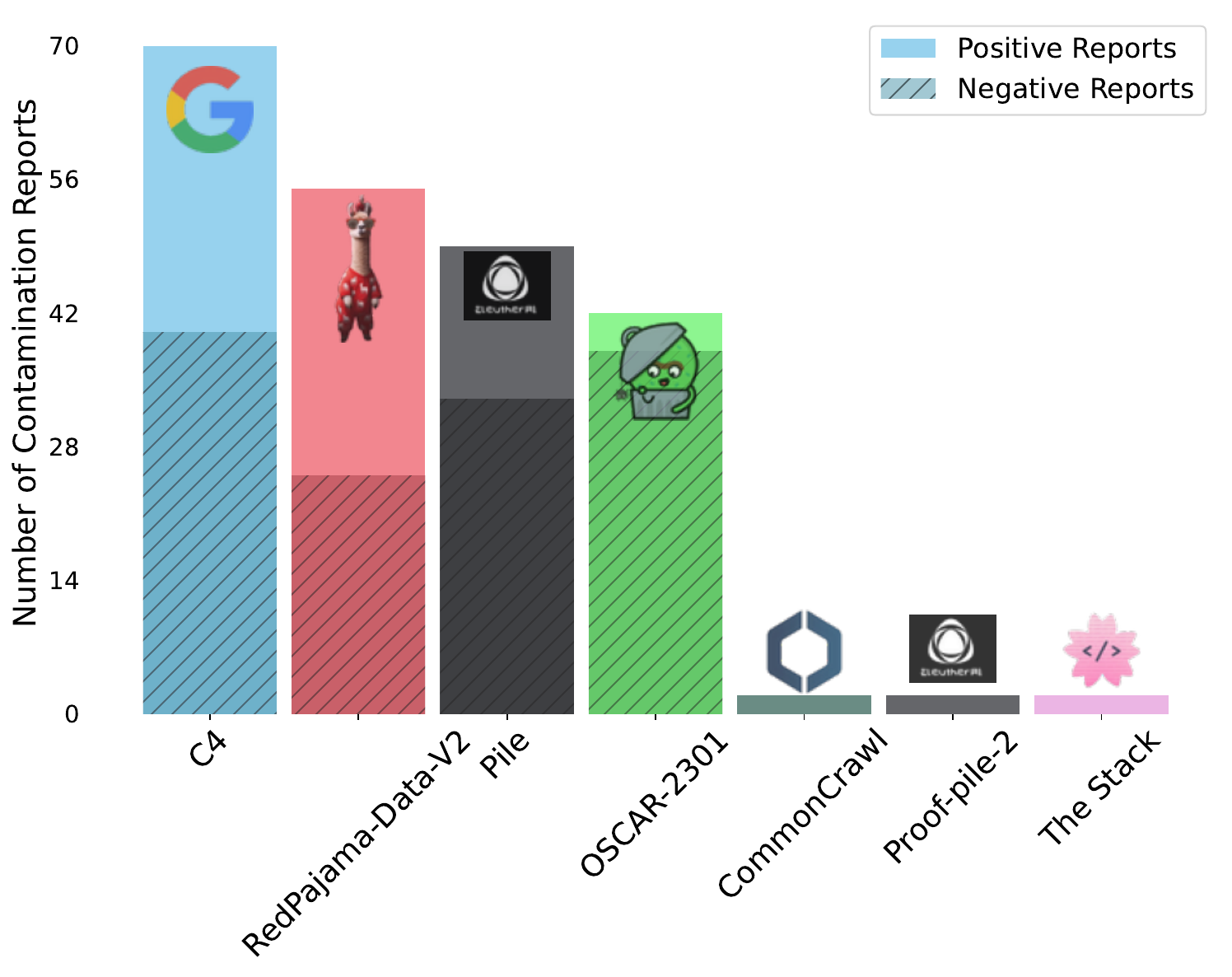}
        \caption{Number of test sets reported for each corpus often used in pre-training.}
        \label{fig:contaminated_corpora}
    \end{minipage}%
    \hfill
    \begin{minipage}[t]{0.48\linewidth}
        \centering
        \includegraphics[width=\linewidth]{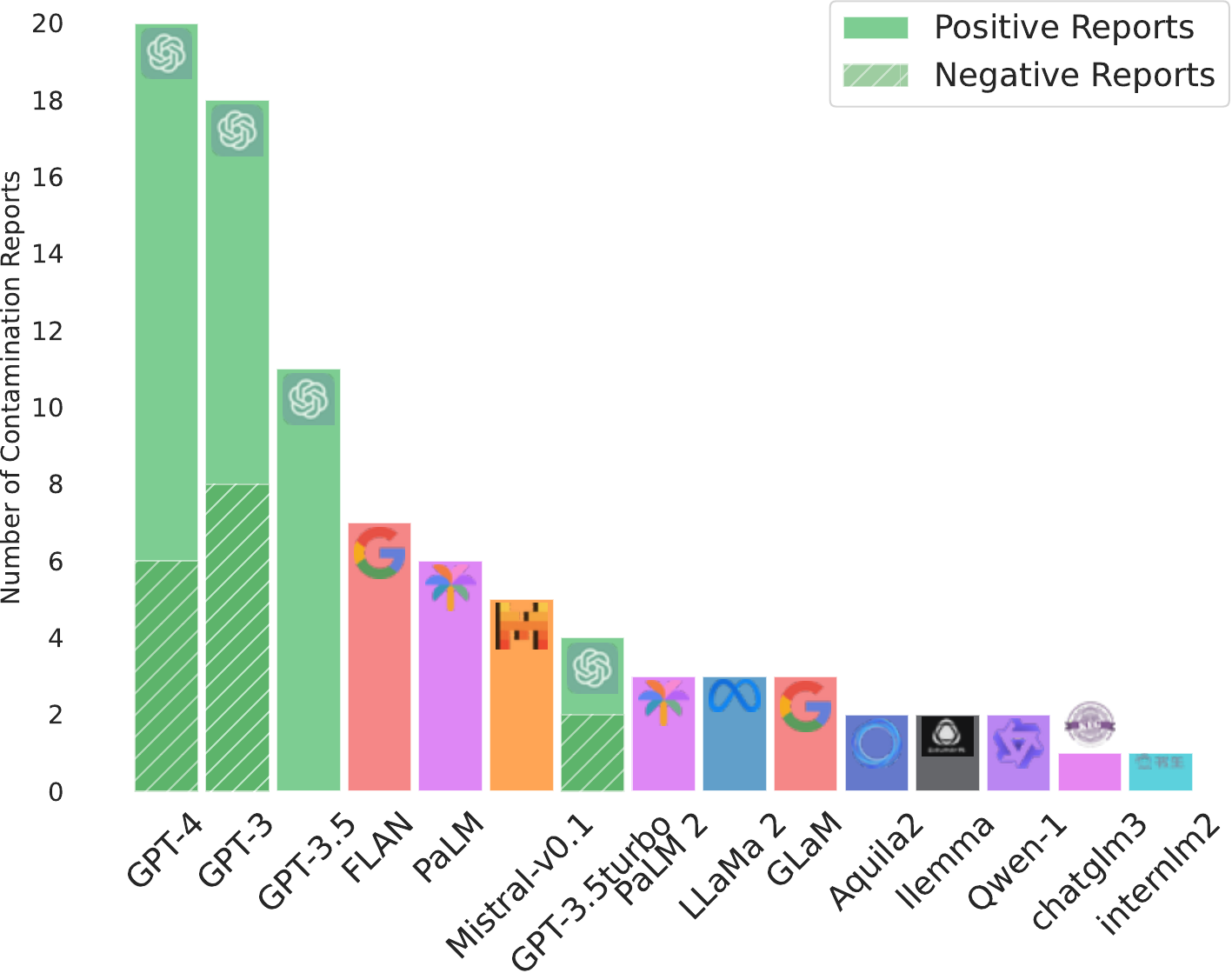}
        \caption{Number of test sets reported for each pre-trained model.}
        \label{fig:contaminated_models}
    \end{minipage}
\end{figure*}

Collecting all the contamination evidence ---or lack of it--- was done openly, through pull requests, and subject to discussions before the admission. Contributors were asked to fill in the information about several aspects, such as the
\textit{contaminated resource} (a training corpus or model), the \textit{evaluation dataset} which was found in the contaminated source, a breakdown of the percentage of contamination found in each split of the dataset (train, development, and test), an optional reference to a paper that describes the methodology behind the submission, as well as whether the contamination detection method was \textit{data-based} or \textit{model-based}. The contributions provided the HuggingFace Hub id of models, corpus, and datasets when possible. In addition, contributors must provide the evidence or a reference to the scientific paper that reported the evidence originally. Figure~\ref{fig:typo-methods} shows the taxonomy of the papers that reported contamination evidence in the shared task.\footnote{Note that there are many other works on data contamination detection. In this report we focus on works that were used to detect contamination for this report. We leave a more detailed coverage survey for future work.} We split the these methods into two: \textit{data-based} and \textit{model-based} approaches.

\paragraph{Data-based approaches.} are methods that inspect the pre-training corpora to find contamination evidence. Data-based approaches typically involve string or sub-string matching techniques such as 13-gram overlap~\cite{brown2020language, wei2022finetuned}, 50-character overlap~\cite{openai2024gpt4} or even full-string overlap~\cite{elazar2024whats}. In Figure~\ref{fig:typo-methods} we differentiate between \textit{Proprietary} and \textit{Open} data. Papers that fall in the category of \textit{Proprietary data} are usually LLMs technical reports that run post-hoc data contamination evaluations to identify and remove evaluation instances that appear in the pre-training corpora~\cite{brown2020language, wei2022finetuned, openai2024gpt4}. Papers that fall in the \textit{open data} category usually involve corpus analysis tools~\cite{dodge-etal-2021-documenting, elazar2024whats} or LLMs with publicly available pre-training data~\cite{azerbayev2023llemma}.

\paragraph{Model-based approaches.} are those methods that try to estimate the contamination of a model by prompting or analyzing the output, without accessing the pre-training data. These methods are formulated as Membership Inference Attacks (MIA) and range from asking LLMs to generate verbatim of the actual evaluation data~\cite{sainz2023chatgpt, golchin2024time} to analyzing the actual output probabilities given by the model~\cite{oren2024proving}. We differentiate between methods applicable to \textit{closed} and \textit{open} models. Methods applicable to \textit{closed} models are usually applicable to \textit{open} models, but not the other way around due to the limitations established by the API or interface providers.

The collected evidences come from different approaches and sources, making them hardly comparable. For transparency, we included in the database information about the source of the evidence and the link to the discussion. We encourage the users to assess how the evidence was collected for their datasets of interest.

\section{Compilation of Evidence} \label{sec:compilation}

The report includes 42 contaminated sources (training corpora or models), 91 datasets, and 566 contamination entries, including 432 contamination events (20 train-set, 95 dev-set, 317 test-set) and 144 non-contamination events, where a contamination event is taken as any report above 0\% of contamination. The database contains, for each split (train, dev, and test) of each evaluation dataset, what percentage was found to be contaminated by a subset of the contamination sources (corpora or models). We analyze separately the contaminated corpora and models.



\begin{figure}[t]
    \centering
        \centering
        \includegraphics[width=\linewidth]{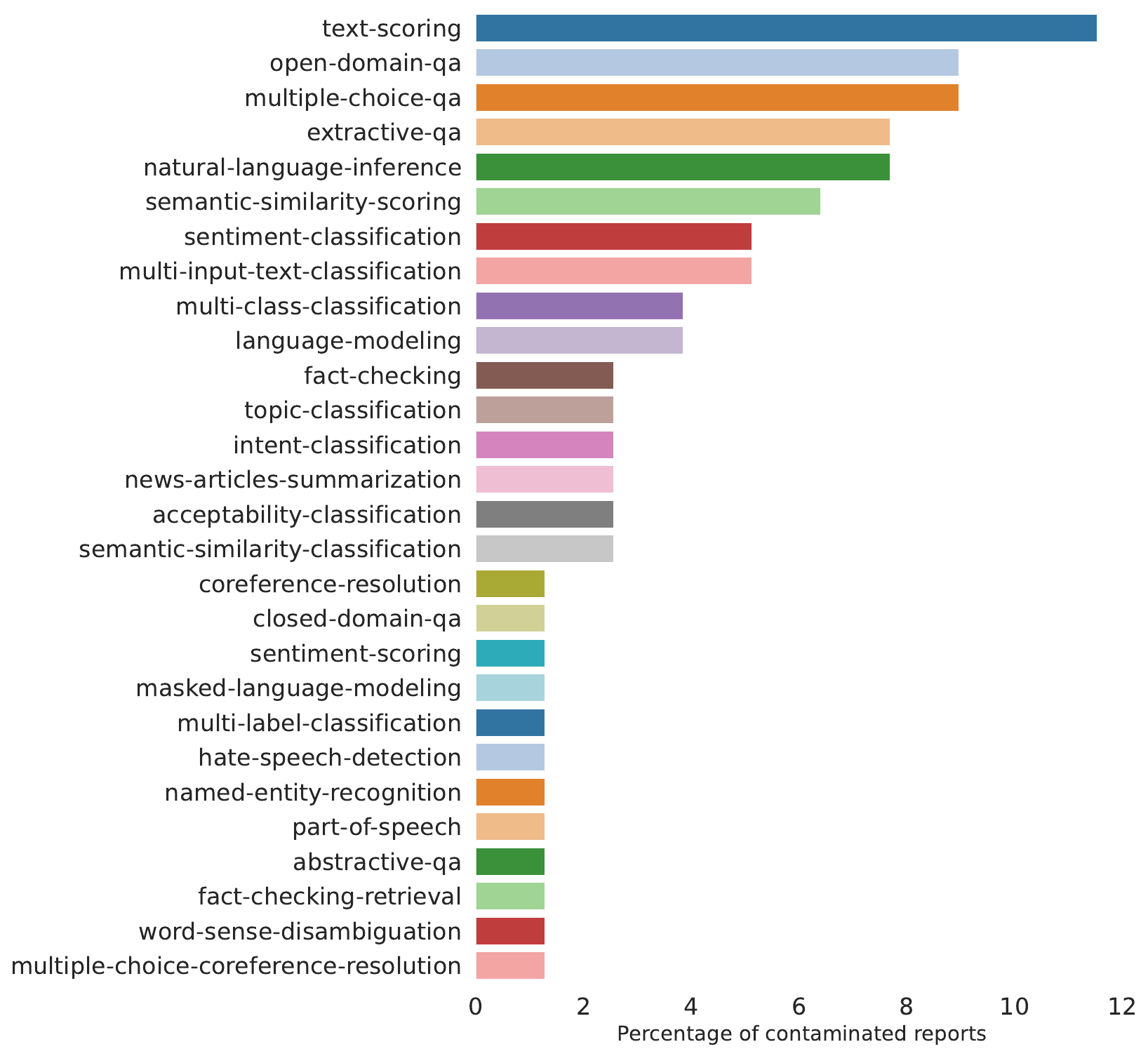}
        \caption{Percentage of contaminated report per task}
        \label{fig:task_report}
\end{figure}

\paragraph{Contaminated corpora.} Figure \ref{fig:contaminated_corpora} shows the number of reported test sets for each corpus often used to pre-train language models. The reported corpora are mainly based on CommonCrawl snapshots, GitHub, or a mix of sources. For CommonCrawl-based corpora, there are 35 events reported for C4~\cite{raffel2023exploring}, 32 for RedPajama v2~\cite{together2023redpajama}, 29 for OSCAR~\cite{2022arXiv221210440J,abadji-etal-2022-towards,AbadjiOrtizSuarezRomaryetal.2021,kreutzer-etal-2022-quality,ortiz-suarez-etal-2020-monolingual,OrtizSuarezSagotRomary2019} and 6 for CommonCrawl~\cite{cc:Rana:2010:Common-Crawl-open-web-scale-crawl} itself. Regarding the GitHub data, there are 2 events reported for the TheStack~\cite{Kocetkov2022TheStack} project. The corpora with various sources, the Pile~\cite{gao2020pile} and ProofPile~\cite{azerbayev2023llemma}, have 30 and 2 reported contamination events respectively. There is also 1 report for xP3~\cite{muennighoff2022crosslingual}, which is a collection of prompts for different NLP datasets.\footnote{The report indicates the use of validation data from a specific dataset as training.}

Table \ref{tab:report-corpora} shows for each corpus often used to pre-train language models, the contamination events involving development or test splits. Please refer to the online database for full details of each report.

\begin{table*}[!ht]\centering
    \setlength\belowcaptionskip{-8px}
    \setlength\tabcolsep{3pt}
    \scriptsize
    \rowcolors{1}{}{lightgray} \ra{1.2}
    \resizebox{0.99\linewidth}{!}{
        \begin{tabular}{
            p{3cm}
            >{\raggedright\arraybackslash}p{10cm}
            }\toprule
            \textbf{Contaminated Source}       & \textbf{Evaluation Set}                                                                                                                                                                                                                                                                                                                                                                               \\\midrule
            allenai/c4 \cite{raffel2023exploring},                        & sem\_eval\_2014\_task\_1 \cite{semeval2014task1}, race, nyu-mll/glue \cite{wang2019glue}, amazon\_reviews\_multi \cite{marc_reviews}, liar \cite{wang2017liar}, reddit\_tifu \cite{kim2018abstractive}, stsb\_multi\_mt \cite{huggingface:dataset:stsb_multi_mt}, wiki\_qa \cite{yang-etal-2015-wikiqa}, gigaword \cite{graff2003english}, piqa \cite{Bisk2020}, esnli \cite{camburu2018esnli}, scitail \cite{scitail}, snli \cite{bowman-etal-2015-large}, ibm/duorc \cite{DuoRC}, math\_qa \cite{amini-etal-2019-mathqa}, swag \cite{zellers2018swagaf}, wiki\_bio \cite{DBLP:journals/corr/LebretGA16}, xnli \cite{conneau2018xnli}, allenai/scicite \cite{cohan2019structural}, aeslc \cite{zhang2019email}, billsum \cite{kornilova2019billsum}, AMR-to-Text, winograd\_wsc \cite{levesque2012winograd}, squadshifts \cite{pmlr-v119-miller20a}, head\_qa \cite{vilares-gomez-rodriguez-2019-head}, xsum \cite{Narayan2018DontGM}, health\_fact \cite{kotonya-toni-2020-explainable}, EdinburghNLP/xsum \cite{Narayan2018DontGM}, UCLNLP/adversarial\_qa \cite{bartolo2020beat}, paws \cite{paws2019naacl}, sick, super\_glue \cite{wang2019superglue}, paws-x \cite{pawsx2019emnlp}, scan, lama \cite{petroni2019language,petroni2020how} \\
            CommonCrawl      \cite{cc:Rana:2010:Common-Crawl-open-web-scale-crawl}                   & allenai/ai2\_arc \cite{allenai:arc}, tau/commonsense\_qa \cite{talmor-etal-2019-commonsenseqa}, ceval/ceval-exam \cite{huang2023ceval}, cais/mmlu \cite{hendryckstest2021}, Rowan/hellaswag \cite{zellers2019hellaswag}, winogrande \cite{levesque2012winograd}                                                                                                                                                                                                                                       \\
            EleutherAI/pile  \cite{gao2020pile}                  & sem\_eval\_2014\_task\_1 \cite{semeval2014task1}, nyu-mll/glue \cite{wang2019glue}, amazon\_reviews\_multi \cite{marc_reviews}, mbpp, openai\_humaneval \cite{chen2021codex}, liar \cite{wang2017liar}, stsb\_multi\_mt \cite{huggingface:dataset:stsb_multi_mt}, wiki\_qa \cite{yang-etal-2015-wikiqa}, gigaword \cite{graff2003english}, piqa \cite{Bisk2020}, esnli \cite{camburu2018esnli}, scitail \cite{scitail}, snli \cite{bowman-etal-2015-large}, ibm/duorc \cite{DuoRC}, swag \cite{zellers2018swagaf}, xnli \cite{conneau2018xnli}, allenai/scicite \cite{cohan2019structural}, aeslc \cite{zhang2019email}, billsum \cite{kornilova2019billsum}, winograd\_wsc \cite{levesque2012winograd}, squadshifts \cite{pmlr-v119-miller20a}, head\_qa \cite{vilares-gomez-rodriguez-2019-head}, xsum \cite{Narayan2018DontGM}, health\_fact \cite{kotonya-toni-2020-explainable}, UCLNLP/adversarial\_qa \cite{bartolo2020beat}, paws \cite{paws2019naacl}, sick, super\_glue \cite{wang2019superglue}, paws-x \cite{pawsx2019emnlp}, scan \\
            oscar-corpus/OSCAR-2301  \cite{2022arXiv221210440J,abadji-etal-2022-towards,AbadjiOrtizSuarezRomaryetal.2021,kreutzer-etal-2022-quality,ortiz-suarez-etal-2020-monolingual,OrtizSuarezSagotRomary2019}          & sem\_eval\_2014\_task\_1 \cite{semeval2014task1}, crows\_pairs \cite{nangia2020crows}, nyu-mll/glue \cite{wang2019glue}, race, amazon\_reviews\_multi \cite{marc_reviews}, openai\_humaneval \cite{chen2021codex}, liar \cite{wang2017liar}, stsb\_multi\_mt \cite{huggingface:dataset:stsb_multi_mt}, wiki\_qa \cite{yang-etal-2015-wikiqa}, gigaword \cite{graff2003english}, piqa \cite{Bisk2020}, esnli \cite{camburu2018esnli}, scitail \cite{scitail}, snli \cite{bowman-etal-2015-large}, math\_qa \cite{amini-etal-2019-mathqa}, swag \cite{zellers2018swagaf}, xnli \cite{conneau2018xnli}, allenai/scicite \cite{cohan2019structural}, aeslc \cite{zhang2019email}, billsum \cite{kornilova2019billsum}, winograd\_wsc \cite{levesque2012winograd}, squadshifts \cite{pmlr-v119-miller20a}, head\_qa \cite{vilares-gomez-rodriguez-2019-head}, xsum \cite{Narayan2018DontGM}, health\_fact \cite{kotonya-toni-2020-explainable}, UCLNLP/adversarial\_qa \cite{bartolo2020beat}, paws \cite{paws2019naacl}, sick, super\_glue \cite{wang2019superglue}                                                        \\
            togethercomputer/RedPajama-Data-V2  \cite{together2023redpajama} & sem\_eval\_2014\_task\_1 \cite{semeval2014task1}, race, nyu-mll/glue \cite{wang2019glue}, amazon\_reviews\_multi \cite{marc_reviews}, liar \cite{wang2017liar}, stsb\_multi\_mt \cite{huggingface:dataset:stsb_multi_mt}, wiki\_qa \cite{yang-etal-2015-wikiqa}, gigaword \cite{graff2003english}, piqa \cite{Bisk2020}, esnli \cite{camburu2018esnli}, scitail \cite{scitail}, snli \cite{bowman-etal-2015-large}, ibm/duorc \cite{DuoRC}, math\_qa \cite{amini-etal-2019-mathqa}, swag \cite{zellers2018swagaf}, xnli \cite{conneau2018xnli}, allenai/scicite \cite{cohan2019structural}, aeslc \cite{zhang2019email}, billsum \cite{kornilova2019billsum}, winograd\_wsc \cite{levesque2012winograd}, squadshifts \cite{pmlr-v119-miller20a}, head\_qa \cite{vilares-gomez-rodriguez-2019-head}, xsum \cite{Narayan2018DontGM}, health\_fact \cite{kotonya-toni-2020-explainable}, UCLNLP/adversarial\_qa \cite{bartolo2020beat}, mc\_taco, paws \cite{paws2019naacl}, samsum \cite{gliwa2019samsum}, sick, super\_glue \cite{wang2019superglue}, paws-x \cite{pawsx2019emnlp}, scan                                              \\
            bigscience/xP3 \cite{muennighoff2022crosslingual}                     & facebook/flores \cite{nllb2022}                                                                                                                                                                                                                                                                                                                                                                                       \\
            EleutherAI/proof-pile-2   \cite{azerbayev2023llemma}        & gsm8k \cite{cobbe2021gsm8k}, hendrycks/competition\_math \cite{hendrycksmath2021}                                                                                                                                                                                                                                                                                                                                                                    \\
            bigcode/the-stack \cite{Kocetkov2022TheStack}                 & openai\_humaneval \cite{chen2021codex}, mbpp                                                                                                                                                                                                                                                                                                                                                                               \\
            \bottomrule
        \end{tabular}
    }
    \caption{A summary of the \textit{dev} or \textit{test} sets found at above 0\% contamination in each corpus often used to pre-train models.}\label{tab:report-corpora}
\end{table*}
\begin{table*}[!t]\centering
    \setlength\belowcaptionskip{-8px}
    \setlength\tabcolsep{3pt}
    \scriptsize
    \rowcolors{1}{}{lightgray} \ra{1.2}
    \resizebox{0.99\linewidth}{!}{
        \begin{tabular}{
            l
            >{\raggedright\arraybackslash}p{9.73cm}
            }\toprule
            \textbf{Contaminated Source} & \textbf{Evaluation Set}                                                                                                                                                                                                                                                                                                                                                                    \\\midrule
            GPT-3 \cite{brown2020language}                       & Reversed Words , race , quac \cite{choi2018quac}, Anagrams 1 , Cycled Letters , mandarjoshi/trivia\_qa \cite{joshi-etal-2017-triviaqa}, ibragim-bad/arc\_easy \cite{allenai:arc}, SAT Analogies, piqa \cite{Bisk2020}, Rowan/hellaswag \cite{zellers2019hellaswag}, wmt/wmt16 \cite{bojar-EtAl:2016:WMT1}, stanfordnlp/coqa \cite{reddy-etal-2019-coqa}, cimec/lambada \cite{paperno-EtAl:2016:P16-1}, natural\_questions \cite{naturalq}, winograd\_wsc \cite{levesque2012winograd}, ucinlp/drop \cite{Dua2019DROP}, rmanluo/RoG-webqsp, rajpurkar/squad\_v2 \cite{rajpurkar-etal-2018-know,rajpurkar-etal-2016-squad}, allenai/openbookqa \cite{Mihaylov2018CanAS}, Symbol Insertion, Anagrams 2 , super\_glue \cite{wang2019superglue}, ibragim-bad/arc\_challenge \cite{allenai:arc}, facebook/anli \cite{nie-etal-2020-adversarial} \\
            GPT-3.5 \cite{brown2020language}                     & samsum \cite{gliwa2019samsum}, yelp\_review\_full \cite{NIPS2015_250cf8b5}, imdb \cite{maas-EtAl:2011:ACL-HLT2011}, ag\_news \cite{Zhang2015CharacterlevelCN}, nyu-mll/glue \cite{wang2019glue}, conll2003 \cite{tjong-kim-sang-de-meulder-2003-introduction}, winogrande \cite{levesque2012winograd}, rajpurkar/squad\_v2 \cite{rajpurkar-etal-2018-know,rajpurkar-etal-2016-squad}, cais/mmlu \cite{hendryckstest2021}, EdinburghNLP/xsum \cite{Narayan2018DontGM}, allenai/openbookqa \cite{Mihaylov2018CanAS}, xlangai/spider \cite{yu-etal-2018-spider}, truthful\_qa \cite{lin2022truthfulqa}                                                                                                                                                                                                       \\
            GPT-4 \cite{openai2024gpt4}                          & samsum \cite{gliwa2019samsum}, yelp\_review\_full \cite{NIPS2015_250cf8b5}, gsm8k \cite{cobbe2021gsm8k}, imdb \cite{maas-EtAl:2011:ACL-HLT2011}, ibragim-bad/arc\_challenge \cite{allenai:arc}, nyu-mll/glue \cite{wang2019glue}, ucinlp/drop \cite{Dua2019DROP}, winogrande \cite{levesque2012winograd}, openai\_humaneval \cite{chen2021codex}, ag\_news \cite{Zhang2015CharacterlevelCN}, EdinburghNLP/xsum \cite{Narayan2018DontGM}, cais/mmlu \cite{hendryckstest2021}, Rowan/hellaswag \cite{zellers2019hellaswag}, allenai/openbookqa \cite{Mihaylov2018CanAS}, truthful\_qa \cite{lin2022truthfulqa}, bigbench \cite{srivastava2023imitation}                                                                                                                                                         \\
            PaLM 2 \cite{anil2023palm}                      & EdinburghNLP/xsum \cite{Narayan2018DontGM}, csebuetnlp/xlsum \cite{hasan-etal-2021-xl}, wiki\_lingua \cite{ladhak-etal-2020-wikilingua}                                                                                                                                                                                                                                                                                                                                          \\
            GPT-3.5-turbo \cite{brown2020language}               & openai\_humaneval \cite{chen2021codex}, HumanEval\_R \cite{chen2021codex}                                                                                                                                                                                                                                                                                                                                                            \\
            FLAN \cite{wei2022finetuned}                         & natural\_questions \cite{naturalq}, mandarjoshi/trivia\_qa \cite{joshi-etal-2017-triviaqa}, story\_cloze \cite{sharma-etal-2018-tackling}, piqa \cite{Bisk2020}, super\_glue \cite{wang2019superglue}, ibragim-bad/arc\_challenge \cite{allenai:arc}, ucinlp/drop \cite{Dua2019DROP}, rajpurkar/squad\_v2 \cite{rajpurkar-etal-2018-know,rajpurkar-etal-2016-squad}, ibragim-bad/arc\_easy \cite{allenai:arc}, Rowan/hellaswag \cite{zellers2019hellaswag}, allenai/openbookqa \cite{Mihaylov2018CanAS}, facebook/anli \cite{nie-etal-2020-adversarial}, winogrande \cite{levesque2012winograd}, wmt/wmt16 \cite{bojar-EtAl:2016:WMT1}                                                                                                                                                \\
            GLaM \cite{du2022glam}                         & stanfordnlp/coqa \cite{reddy-etal-2019-coqa}, natural\_questions \cite{naturalq}, mandarjoshi/trivia\_qa \cite{joshi-etal-2017-triviaqa}, story\_cloze \cite{sharma-etal-2018-tackling}, cimec/lambada \cite{paperno-EtAl:2016:P16-1}, piqa \cite{Bisk2020}, super\_glue \cite{wang2019superglue}, ibragim-bad/arc\_challenge \cite{allenai:arc}, race , quac \cite{choi2018quac}, winograd\_wsc \cite{levesque2012winograd}, rajpurkar/squad\_v2 \cite{rajpurkar-etal-2018-know,rajpurkar-etal-2016-squad}, ibragim-bad/arc\_easy \cite{allenai:arc}, Rowan/hellaswag \cite{zellers2019hellaswag}, allenai/openbookqa \cite{Mihaylov2018CanAS}, facebook/anli \cite{nie-etal-2020-adversarial}, winogrande \cite{levesque2012winograd}                                                                                                            \\
            LLaMa 2-13B \cite{touvron2023llama}                  & allenai/openbookqa \cite{Mihaylov2018CanAS}, winogrande \cite{levesque2012winograd}, truthful\_qa \cite{lin2022truthfulqa}                                                                                                                                                                                                                                                                                                                                               \\
            Mistral-7B \cite{jiang2023mistral}                   & allenai/openbookqa \cite{Mihaylov2018CanAS}, winogrande \cite{levesque2012winograd}, truthful\_qa \cite{lin2022truthfulqa}, cais/mmlu \cite{hendryckstest2021}                                                                                                                                                                                                                                                                                                                                    \\
            PaLM \cite{chowdhery2022palm}                        & cimec/lambada \cite{paperno-EtAl:2016:P16-1}, super\_glue \cite{wang2019superglue}, ibragim-bad/arc\_challenge \cite{allenai:arc}, winograd\_wsc \cite{levesque2012winograd}, rmanluo/RoG-webqsp, rajpurkar/squad\_v2 \cite{rajpurkar-etal-2018-know,rajpurkar-etal-2016-squad}, mandarjoshi/trivia\_qa \cite{joshi-etal-2017-triviaqa}, ibragim-bad/arc\_easy \cite{allenai:arc}                                                                                                                                                                                                                              \\
            Claude 3 Opus               & facebook/flores \cite{nllb2022}                                                                                                                                                                                                                                                                                                                                                                            \\
            bigscience/bloomz \cite{muennighoff2022crosslingual}           & facebook/flores \cite{nllb2022}                                                                                                                                                                                                                                                                                                                                                                            \\
            bigscience/mt0-* \cite{muennighoff2022crosslingual}        & facebook/flores \cite{nllb2022}                                                                                                                                                                                                                                                                                                                                                                            \\
            BAAI/Aquila2-34B             & gsm8k \cite{cobbe2021gsm8k}, hendrycks/competition\_math \cite{hendrycksmath2021}                                                                                                                                                                                                                                                                                                                                                         \\
            BAAI/AquilaChat2-34B         & gsm8k \cite{cobbe2021gsm8k}                                                                                                                                                                                                                                                                                                                                                                                      \\
            EleutherAI/llemma\_* \cite{azerbayev2023llemma}       & gsm8k \cite{cobbe2021gsm8k}, hendrycks/competition\_math \cite{hendrycksmath2021}                                                                                                                                                                                                                                                                                                                                                         \\
            Qwen/Qwen-1\_8B \cite{qwen}              & gsm8k \cite{cobbe2021gsm8k}, hendrycks/competition\_math \cite{hendrycksmath2021}                                                                                                                                                                                                                                                                                                                                                         \\
            BAAI/Aquila2-7B              & hendrycks/competition\_math \cite{hendrycksmath2021}                                                                                                                                                                                                                                                                                                                                                                \\
            Qwen/Qwen-* \cite{qwen}                & hendrycks/competition\_math \cite{hendrycksmath2021}                                                                                                                                                                                                                                                                                                                                                                \\                                                                         
            THUDM/chatglm3-6b \cite{du2022glm}           & hendrycks/competition\_math \cite{hendrycksmath2021}                                                                                                                                                                                                                                                                                                                                                                \\
            internlm/internlm2-*  \cite{cai2024internlm2}         & hendrycks/competition\_math \cite{hendrycksmath2021}                                                                                                                                                                                                                                                                                                                                                                \\
            mistralai/Mistral-7B-v0.1 \cite{jiang2023mistral}    & ibragim-bad/arc\_easy \cite{allenai:arc}                                                                                                                                                                                                                                                                                                                                                                      \\
            \bottomrule
        \end{tabular}
    }
    \caption{A summary of the \textit{dev} or \textit{test} sets found at above 0\% contamination in each reported model. The "*" is used to indicate the different versions or sizes of the models.}
    \label{tab:report-models}
\end{table*}


\paragraph{Contaminated models.} Figure \ref{fig:contaminated_models} details the number of contamination events involving test sets that were reported, organised according to each pre-trained model. Most reported evidence is for closed models, for instance: 24 for GPT-3~\cite{brown2020language}, 17 for GLaM~\cite{du2022glam}, 16 for GPT-4~\cite{openai2024gpt4}, 13 for GPT-3.5~\cite{brown2020language}, 8 for PaLM~\cite{chowdhery2022palm}, 3 for PaLM-2~\cite{anil2023palm}, 2 for GPT-3.5 Turbo~\cite{brown2020language} and 1 for Calude 3 Opus. In the case of open models: there are 14 reported events for models fine-tuned with FLAN data~\cite{wei2022finetuned}, 5 for Mistral~\cite{jiang2023mistral}, 3 for Llama 2~\cite{touvron2023llama}, 2 for Qwen~\cite{qwen}, Llema~\cite{azerbayev2023llemma} and Aquila 2; and a single one for mT0 and Bloom-Z~\cite{muennighoff2022crosslingual}.

Table \ref{tab:report-models} shows for each pre-trained language model, the contamination events involving development or test splits. Please refer to the online database for full details of each report.

\section{Analysis of the Reported Data} \label{sec:trends}

In this section, we analyze the reported entries to understand the report's data better. 

\begin{figure}[t]
    \centering
    \includegraphics[width=\linewidth]{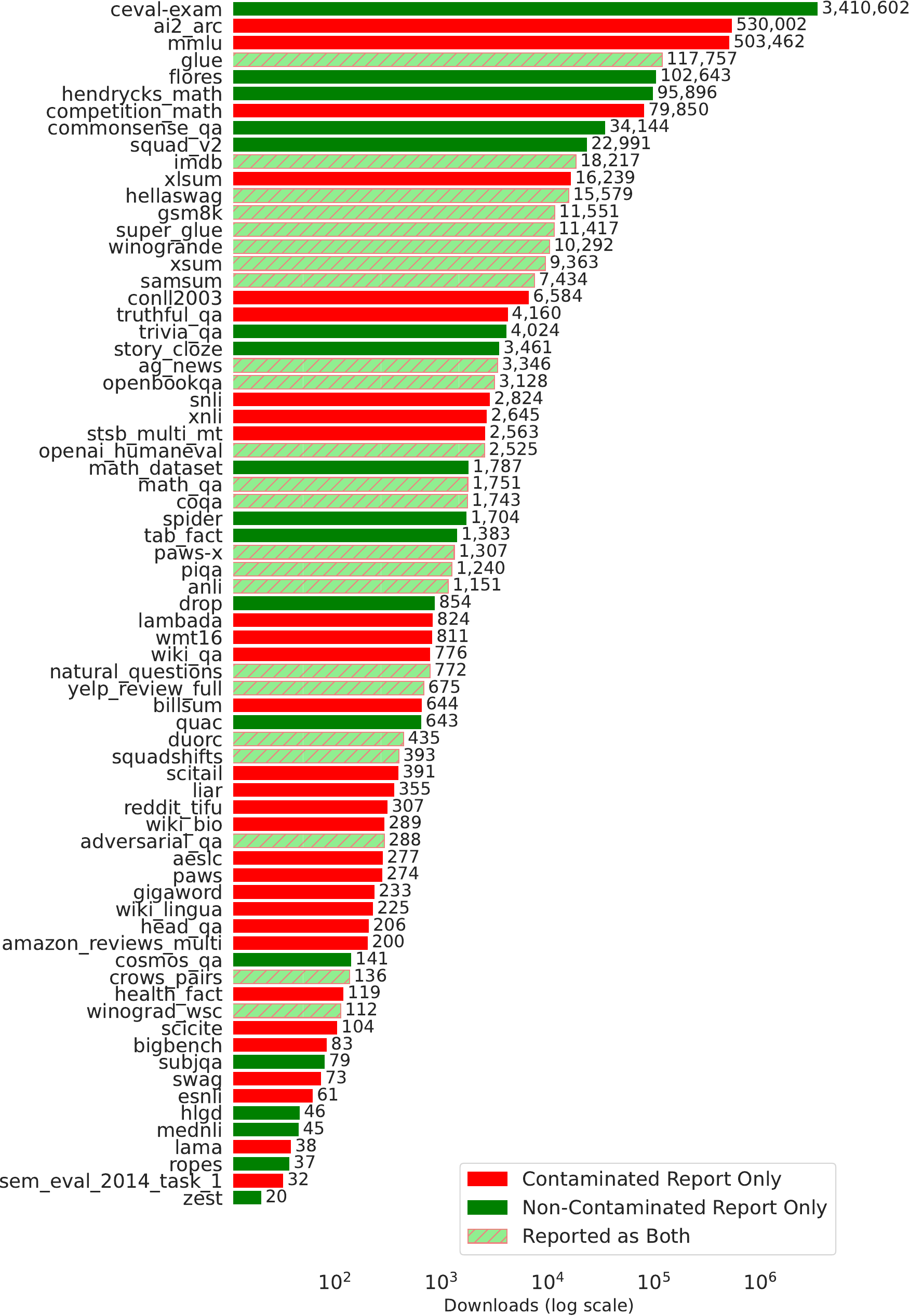}
    \caption{Number of downloads in the HuggingFace hub of the datasets in the report.}
    \label{fig:downloads}
\end{figure}

\paragraph{Reported tasks.} Figure \ref{fig:task_report} shows the percentage of data contamination per task. We use the \verb+task_id+ assigned to each dataset in the Hugging Face hub. Text-scoring, QA, and multiple-choice-qa are among the most contaminated task types. Figure ~\ref{fig:downloads} shows the number of downloads for every dataset in the report. We measure the total number of downloads from the Hugging Face hub.\footnote{\url{https://huggingface.co/docs/datasets}} Since one model may be reported as contaminated with a dataset while another model may not, we have entries of both being compromised and non-compromised for some datasets. Relating both tables, we can see that the tasks reported as the most contaminated include very popular datasets such as MMLU (multiple-choice-qa), GLUE (text-scoring), and ai2\_arc (multiple-choice-qa), which are standard benchmarks for measuring the performance of LLMs. These benchmarks, as well as other very popular benchmarks reported in instances of data contamination, such as hellaswag or gsm8k are implemented in community leaderboards such as the Open LLM Leaderboard.\footnote{\url{https://hf.co/spaces/open-llm-leaderboard/}}

\begin{figure*}[t]
    \centering
    \begin{minipage}[t]{0.48\linewidth}
        \centering
\includegraphics[width=\linewidth]{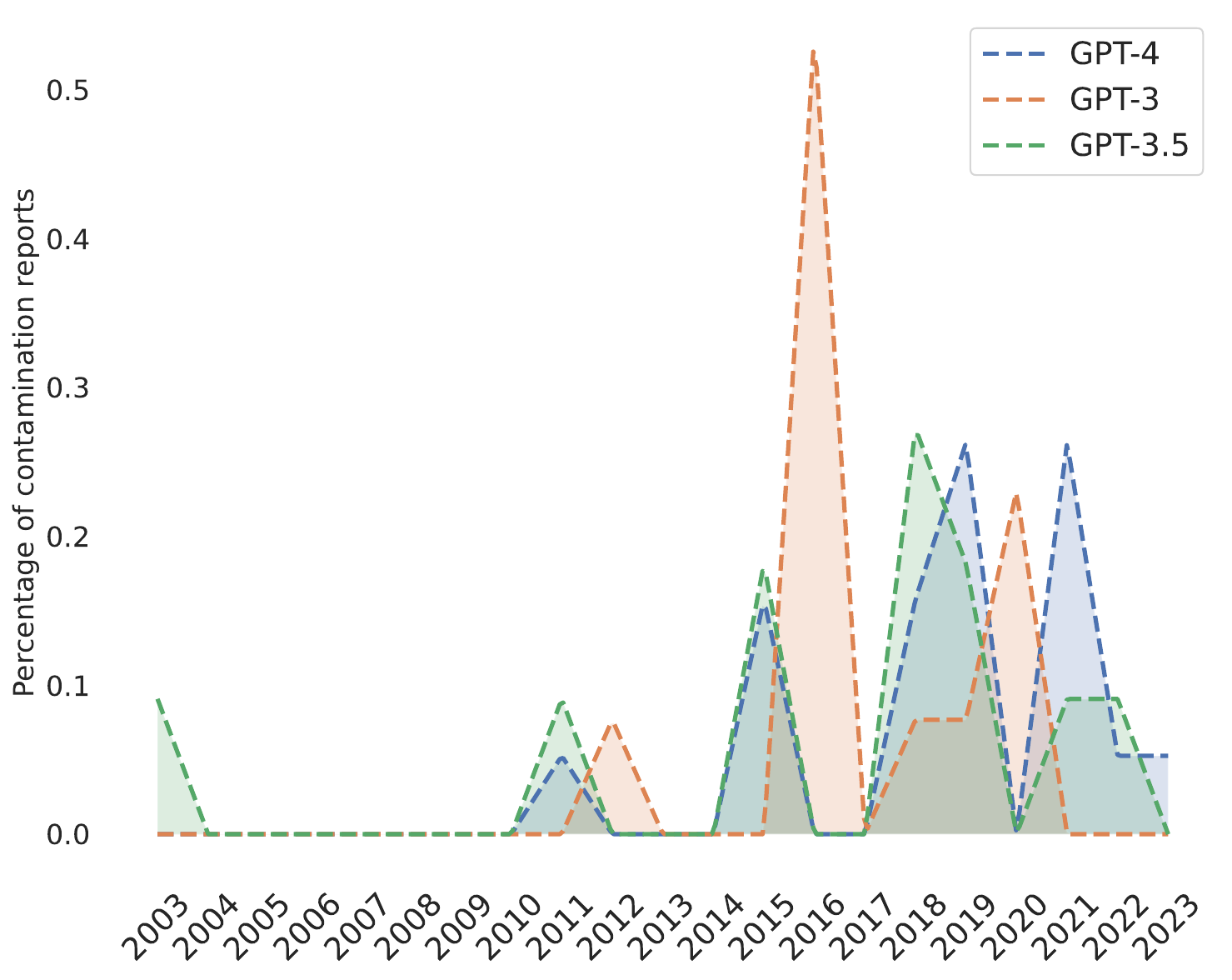}
    \caption{Year of publication of the contaminated test sets reported for each model.}
    \label{fig:contamion_year_model}
    \end{minipage}%
    \hfill
    \begin{minipage}[t]{0.48\linewidth}
        \centering
        \includegraphics[width=\linewidth]{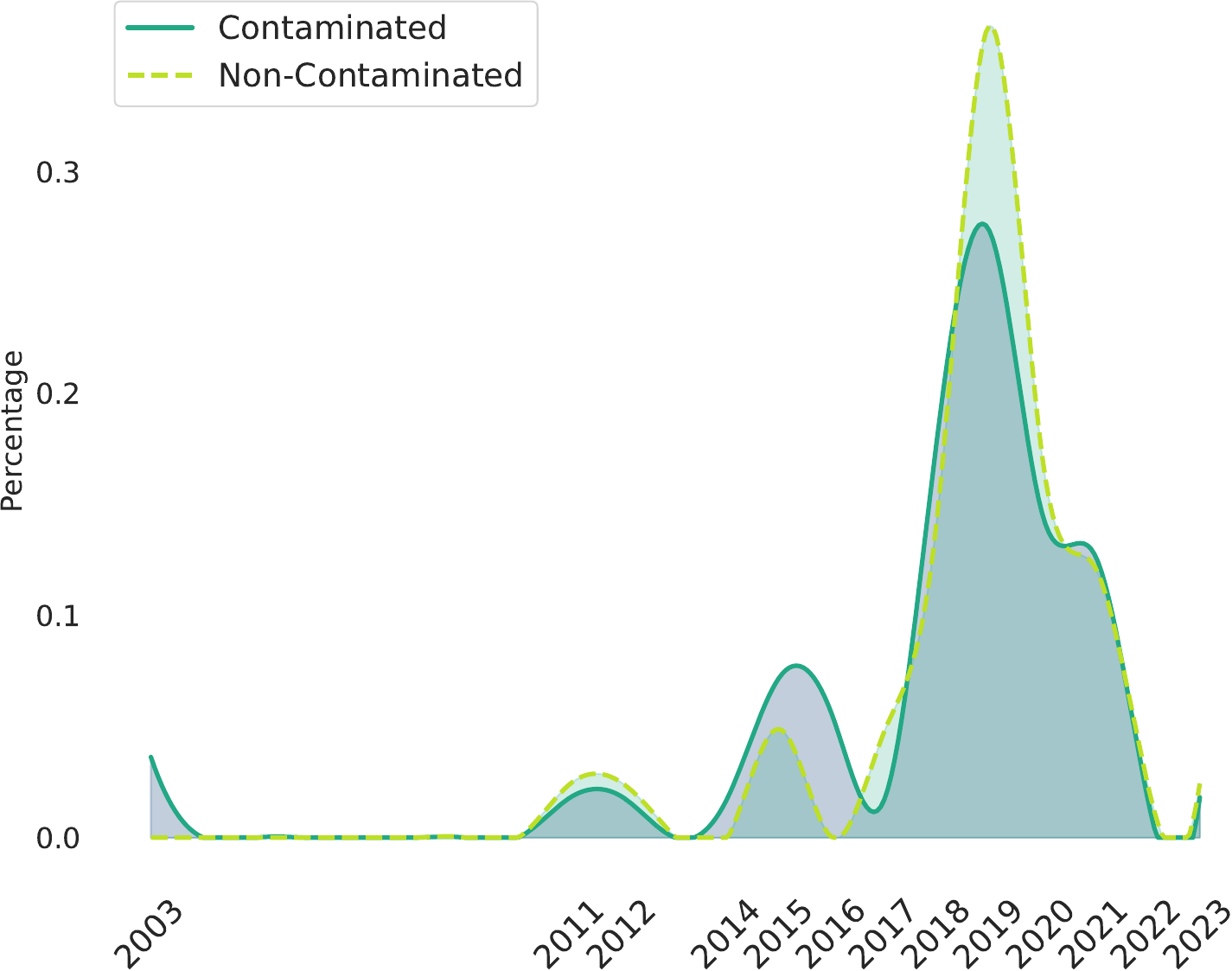}
        \caption{Publication year of the test sets included in the data contamination report.}
        \label{fig:contamion_year}
    \end{minipage}
\end{figure*}

\paragraph{Year of publication of the reported data.} Figure \ref{fig:contamion_year} shows the percentage of total test sets included in contamination events per year. We present data for test sets in both contamination events (>0\% contamination) and non-contamination events (0\% contamination). Most of the reported datasets correspond to the 2018 to 2021 period. 

We further explore the relationship between the year of publication of the datasets and instances of contamination by examining the reported data contamination for the three models with the most instances of data contamination: GPT-4, GPT-3, and GPT-3.5. As expected based on the models' release dates, Figure \ref{fig:contamion_year_model} shows that more recently released models are contaminated with more recently released datasets. For instance, GPT-3, launched in 2020, is predominantly contaminated with datasets from 2016, while GPT-4, released in 2023, is mainly contaminated with datasets from 2018 to 2022. 




\section{Conclusions}

Data contamination has become a significant concern in recent times. Consequently, a growing number of papers and state-of-the-art models mention issues of data contamination. In the CONDA 2024 Shared Task on Evidence of Data Contamination, we have collected and compiled a comprehensive database of available evidence on data contamination in currently available datasets and models. This report includes 566 contamination entries over 91 contaminated sources from a total of 23 contributors. With this shared task, we provide a structured, centralized platform for contamination evidence collection to help the community understand the extent of the problem and to assist researchers in avoiding reporting evaluation results on known contaminated resources. Given the large exploration space, this report does not cover all cases, but a small sample that were reported during our shared task period, in the midst of 2024. We welcome further submissions to the database, and plan to keep this database up-to-date as it provides a valuable source of information for the research community.

\section*{Acknowledgments}

We are grateful to Hugging Face and Clémentine Fourrier for their support in establishing the website for the Data Contamination Database hosted on Hugging Face Spaces.
We acknowledge the support of project Disargue (TED2021-130810B-C21, funded by MCIN/AEI /10.13039/501100011033 and by European Union NextGenerationEU/ PRTR) and the Basque Government (Research group funding IT-1805-22). 

\bibliography{custom}
\clearpage




\end{document}